\documentclass[letterpaper, 10 pt, conference]{ieeeconf}  

\IEEEoverridecommandlockouts                              

\usepackage{amsmath,amsfonts}
\usepackage{algorithmic}
\usepackage{algorithm}
\usepackage{caption}
\usepackage{booktabs}
\usepackage{multirow}
\usepackage{makecell}
\usepackage{diagbox}
\usepackage{array}
\usepackage{booktabs}
\usepackage{bigstrut}
\usepackage{color}
\usepackage{xcolor}
\usepackage{colortbl}
\usepackage[colorlinks,
            linkcolor=green,
            anchorcolor=green,
            citecolor=green]{hyperref}
\definecolor{mypink}{rgb}{.75,1,.75}
            
\usepackage{authblk}
\usepackage{array}
\usepackage[caption=false,font=normalsize,labelfont=sf,textfont=sf]{subfig}
\usepackage{textcomp}
\usepackage{stfloats}
\usepackage{url}
\usepackage{verbatim}
\usepackage{graphicx}
\usepackage{cite}
\usepackage{bbm}
\definecolor{mygreen}{rgb}{.75,1,.75}
\usepackage{etoolbox}
\makeatletter
\patchcmd{\@makecaption}
{\scshape}
{}
{}
{}

\usepackage{pifont}

\newcommand{\xmark}{\ding{55}}
\usepackage[capitalize]{cleveref}
\makeatother

\title{\LARGE \bf
RockTrack: A 3D Robust Multi-Camera-Ken \\
Multi-Object Tracking Framework
}

\author{Xiaoyu Li$^{1, 2,^\dagger}$, Peidong Li$^{2,^\dagger}$, Lijun Zhao$^{1}$, Dedong Liu$^{1}$, Jinghan Gao$^{1}$, Xian Wu$^{1}$, Yitao Wu$^{1}$, Dixiao Cui$^{2}$
	\thanks{This work was done while Xiaoyu Li was interning at Zhijia Technology.}
	\thanks{Corresponding author: Lijun Zhao and Dixiao Cui. $^\dagger$ : These authors contributed equally to this work.}
        \thanks{$^1$State Key Laboratory of Robotics and Systems, Harbin Institute of Technology, China, $^2$Zhijia Technology, China}
        }

\begin{document}

\maketitle
\thispagestyle{empty}
\pagestyle{empty}

\begin{abstract}

3D Multi-Object Tracking (MOT) obtains significant performance improvements with the rapid advancements in 3D object detection, particularly in cost-effective multi-camera setups. 
However, the prevalent end-to-end training approach for multi-camera trackers results in detector-specific models, limiting their versatility.
Moreover, current generic trackers overlook the unique features of multi-camera detectors, i.e., the unreliability of motion observations and the feasibility of visual information.
To address these challenges, we propose RockTrack, a 3D MOT method for multi-camera detectors. 
Following the Tracking-By-Detection framework, RockTrack is compatible with various off-the-shelf detectors.
RockTrack incorporates a confidence-guided preprocessing module to extract reliable motion and image observations from distinct representation spaces from a single detector. 
These observations are then fused in an association module that leverages geometric and appearance cues to minimize mismatches. 
The resulting matches are propagated through a staged estimation process, forming the basis for heuristic noise modeling. 
Additionally, we introduce a novel appearance similarity metric for explicitly characterizing object affinities in multi-camera settings.
RockTrack achieves state-of-the-art performance on the nuScenes vision-only tracking leaderboard with 59.1\% AMOTA while demonstrating impressive computational efficiency.

\end{abstract}

\section{INTRODUCTION}

\begin{figure}[t]
      \centering
      \includegraphics[width=\linewidth]{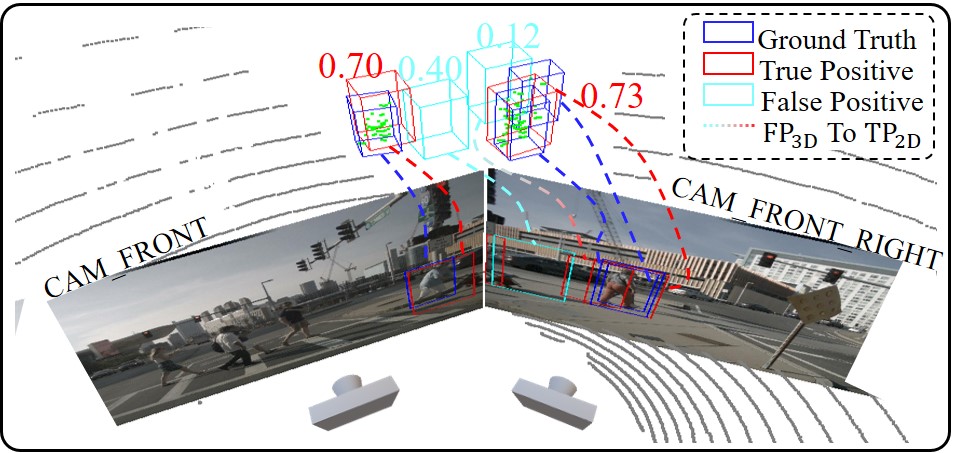}
      \vspace{-1em}
      \caption[l]{The visualization of HeightTrans~\cite{li2024dualbev} highlights specific features of multi-view detectors. 
      (\uppercase\expandafter{\romannumeral1}) \textit{The inherent ill-posed nature} results in unreliable 3D estimations, i.e., 0.4 score $FP_{3D}$.
      (\uppercase\expandafter{\romannumeral2}) \textit{The interaction across various representation spaces} allows low-score detections, i.e., 0.12 score $FP_{3D}$, to be valid visual residuals for certain tracklets.
      }
      \vspace{-2em}
    \label{fig:f0}
   \end{figure}

    3D Multi-Object Tracking~\cite{ab3dmot, polymot, centerpoint, simpletrack, camomot, eagermot, shasta, deepfusionmot, cc3dt, Monotrack, mutr3d, li2024fastpoly, ding2024ada, lin2023sparse4d} (MOT) provides the smooth and abundant motion information of surrounding obstacles. 
    Rapid advancements in 3D detection have significantly enhanced 3D MOT.
    Multi-view detectors~\cite{lss, bevdet, DETR3D, BEVFormer, PETR, streampetr} are gradually emerging as a predominant 3D detection paradigm with their cost-effectiveness and promising performance.
    Multi-camera MOT approaches~\cite{ding2024ada, mutr3d, pftrack, li2023end, lin2023sparse4d} typically leverage shared high-level features extracted by the detection network, which constrains their adaptability across diverse detector architectures.
    Furthermore, generic 3D trackers~\cite{ab3dmot, polymot, simpletrack, li2024fastpoly} adhering to the \textit{Tracking-By-Detection (TBD)} framework are primarily designed for LiDAR-based detectors, resulting in limited support for camera-only systems.
    As illustrated in Fig.~\ref{fig:f0}, two key characteristics of these detectors warrant further investigation for 3D MOT:
    
    (\uppercase\expandafter{\romannumeral1}) \textit{\textbf{Con}: The ill-posed nature of extracting spatial information from 2D images leads to unreliable 3D estimations.} 
    To recall valuable detections from raw noisy detections, existing 3D trackers typically employ Non-Maximum Suppression (NMS)~\cite{polymot, camomot, cc3dt, simpletrack, li2024fastpoly} and Score Filter (SF)~\cite{camomot, polymot, li2024fastpoly}.
    However, due to the uncertainty of depth estimation, multi-camera detectors generate numerous low-confidence False Positive (FP). 
    This necessitates the tracker to apply stringent thresholds, achieving a recall-precision trade-off and attaining the optimal tracking performance, as shown in Fig. \ref{fig:f5}(c).
    A concern arises that low-score but valid detections (due to illuminate change, etc.) may fail to recall.
    Additionally, as illustrated in~\cite{bevdet} and Fig.~\ref{fig:f0}, the unreliability even results in no intersection between high-score FP and True Positive (TP), deactivating the traditional NMS. 
    
    (\uppercase\expandafter{\romannumeral2}) \textit{\textbf{Pro}: The mutual transformation between camera features and structured features enables these detectors to convey image (2D) and spatial (3D) information.}
    Specifically, multi-camera detectors extract 3D object information from images, either explicitly~\cite{DETR3D, PETR, streampetr} or via intermediate variables (Bird's Eye View (BEV) feature)~\cite{li2024dualbev, BEVFormer, solofusion}.
    This architecture results in 3D detection outputs encompassing potentially valid 2D information, as displayed in Fig.~\ref{fig:f0}.
    However, current trackers fail to effectively utilize visual observations from these detectors, typically relying on additional 2D detectors to fuse appearance information.
    Nevertheless, they suffer from spatio-temporal alignment across diverse modalities, particularly in multi-camera settings.
    
    Toward this end, we present a 3D \textbf{Ro}bust Multi-\textbf{C}amera-View(\textbf{K}en) \textbf{Track}er tailored for the camera-only detectors, termed \textbf{RockTrack}.    
    To ensure generalization, RockTrack adheres to the TBD framework and introduces a novel multi-camera appearance similarity metric (MCAS) that can be integrated into various trackers.
    In response to the characteristics of multi-camera detectors, with a single detector as input, RockTrack employs a confidence-guided pre-processing module to maximize the recall of reliable observations from distinct embedding spaces.
    Specifically, to counter inaccurate depth estimations, we propose a scalable geometry filter to eliminate non-overlapping 3D FP in high-score detections.
    Leveraging potentially viable 2D information, we introduce a pseudo visual tracker filter to selectively recover valid low-confidence detections.
    We utilize a multi-modal data association to optimize the utilization of appearance and spatial information.
    The matching results are progressively propagated to the estimation module to model the motion noise, further enhancing the tracking robustness to unreliable observations.
    RockTrack builds upon our previous work, Poly-MOT~\cite{polymot}.
    As an unsupervised method, RockTrack achieves \textbf{state-of-the-art performance on the nuScenes vision-only tracking leaderboard with 59.1\% AMOTA}, while maintaining competitive runtime performance using only CPU.
    The main contributions of this work include:
\begin{itemize}
        \item We present RockTrack, a robust and flexible 3D MOT method based on the TBD framework, specifically designed for multi-view detectors.
        \item We introduce a novel multi-view appearance similarity metric to capture inter-object affinity explicitly.
        \item We develop a scalable geometry filter and adapt motion measurement noise to enhance spatial reliability in multi-camera detectors. We also propose a pseudo visual tracker filter and implement multi-modal matching to leverage visual information effectively.
        \item RockTrack establishes a new state-of-the-art with 59.1\% AMOTA on the nuScenes vision-only test leaderboard.
\end{itemize}

\section{Related Work}

\textbf{Camera-only 3D Detection.}
Earlier works like FCOS3D~\cite{fcos3d} improve depth modeling with additional 3D regression branches in the monocular image detector.
Multi-camera fusion leads camera-only 3D detection due to its clear spatio-temporal representation.
\cite{bevdet, lss, solofusion} generate discrete depth values for individual pixels, projecting them into 3D space and converting them into BEV space.
Correspondingly, \cite{BEVFormer, li2024dualbev} implement 3D-to-2D cross-attention based on the transformer~\cite{BEVFormer} or CNN~\cite{li2024dualbev} to facilitate the interaction of different embedding spaces.
Besides, \cite{DETR3D, PETR, streampetr, lin2023sparse4d} decode object-structured information directly from images through sparse query operations.
However, these detectors inherit an ill-posed nature due to the inherent lack of depth information, limiting accuracy and compromising subsequent tracking tasks.
RockTrack leverages a confidence-guided pre-processing and an adaptive noise module to maximize reliable 3D observation recall from challenging raw detections.

\textbf{Camera-only 3D MOT.}
Camera-only 3D MOT methods~\cite{cc3dt, Monotrack, mutr3d, pftrack, li2023end, huang2023delving, ding2024ada} are gaining attention with the growing efficiency of camera-only 3D detectors.
\textcolor{black}{Early methods~\cite{cc3dt, Monotrack, huang2023delving} rely on the monocular detector and the TBD framework to accomplish cross-camera, cross-object tracking.
They collect and aggregate the trajectory~\cite{Monotrack} or detection~\cite{cc3dt} results from all individual cameras.
\cite{huang2023delving} employs a spatio-temporal transformer to the motion information of objects, followed by a learning-based matching process.
However, these trackers are limited by monocular detection performance and lack interaction between cameras.}
Recent trackers~\cite{mutr3d, pftrack, li2023end, ding2024ada, lin2023sparse4d}, benefiting from advanced multi-camera detectors, implicitly learn refinement of tracklets, achieving exceptional temporal modeling.
\cite{mutr3d} pioneers a framework based on 3D queries~\cite{DETR3D} for coherent spatial and appearance tracks.
\cite{pftrack} employs attention-based architecture for tracklet propagation, enhancing the robustness of the tracker.
\cite{li2023end, ding2024ada} employ task-specific queries to decouple tracking and detection within an end-to-end framework.
However, these trackers jointly train the detection network, rendering them detector-specific.
In contrast, RockTrack follows the TBD framework, can be extended to various detectors, and ensures superior tracking results.

\begin{figure*}[t]
\centering
\includegraphics[width=1\textwidth]{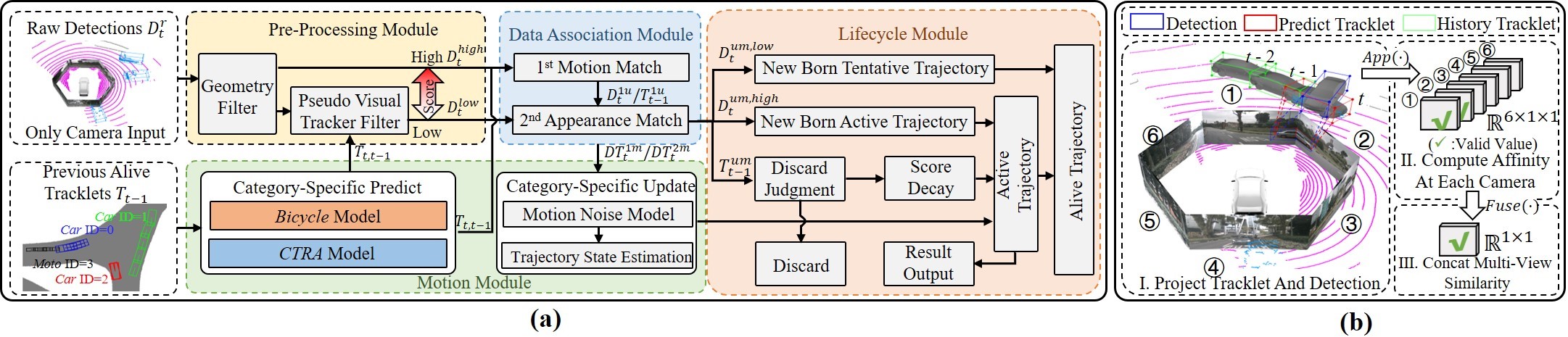}
\vspace{-1.5em}
\caption[l]{\textbf{(a): The pipeline of RockTrack. }
\textcolor{black}{
(\uppercase\expandafter{\romannumeral1}): Previous tracklets $T_{t-1}$ are predicted to $T_{t, t-1}$.
(\uppercase\expandafter{\romannumeral2}): 3D raw detections $D^{r}_{t}$ undergo a geometry filter to reduce FP, resulting in $D_{t}$.
A visual filter is then introduced to recall valid low-score detections in $D_{t}$, obtaining $D^{low}_{t}$.
(\uppercase\expandafter{\romannumeral3}): High-score detections $D^{high}_{t}$ in $D_{t}$ are first matched with $T_{t, t-1}$ based on motion.
$D^{low}_{t}$, unmatched detections in $D^{high}_{t}$ and tracklets, are then associated through appearance.
(\uppercase\expandafter{\romannumeral4}): Final matched detections and heuristic observation-specific noises are utilized to update the corresponding tracklets.
(\uppercase\expandafter{\romannumeral5}):
The count-based lifecycle module is employed to init, penalize, and discard tracklets.
All alive tracklets merge to obtain $T_{t}$ and forward to the subsequent frame.}
\textbf{(b): The calculation process of our proposed multi-camera appearance similarity metric (MCAS). ($K=6$)}
}
\vspace{-2em}
\label{fig:f2}
\end{figure*}

\section{RockTrack}
\textcolor{black}{Our method consists of four parts: pre-processing module, data association module, motion module, and lifecycle module, as illustrated in Fig.~\ref{fig:f2}(a).}

\subsection{Confidence-Guided Pre-Processing Module}
        \label{Pre-processing}
        As highlighted in Fig. \ref{fig:f0}, within camera-only detectors, even the high-score detections contain considerable FP without intersection with TP, deactivating the classic NMS in current 3D trackers.
        In addition, low-confidence detections might contain effective visual information, posing a recall-failed challenge to SF.
        Toward these, guided by detection confidence, we utilize specialized filters to selectively recall pertinent observations in 3D/BEV space and 2D space. 
        \textcolor{black}{Generally, a scalable geometry-based filter is first applied to all detections $D^{r}_{t}$ ($r$ for raw) to suppress 3D FP.
        This process is expressed as:}
        \begin{equation}
            \textcolor{black}{D_{t} = \text{Geo-Filter}(D^{r}_{t}),}
            \label{eq:geometry_filter}
        \end{equation}
        \textcolor{black}{where $D_{t}$ is the processed detections.
        $\text{Geo-Filter}(\cdot)$ is our proposed geometry filter.
        Subsequently, we utilize hand-crafted thresholds $\theta_{sf}$ ($sf$ for score filter) to divide high/low-confidence detections in $D_{t}$, obtaining $D^{high}_{t}$ and $D^{low, c}_{t}$ ($c$ for coarse).
        $D^{high}_{t}$ are retained directly since they are relatively reliable.
        Instead of directly discarding~\cite{polymot, camomot} or retaining~\cite{simpletrack, bytetrack} $D^{low, c}_{t}$, we leverage a pseudo visual tracker to excavate valuable detections based on their appearance cue with existing tracklets.
        This process is described as:}
        \begin{equation}
            \textcolor{black}{D^{low}_{t} = \text{Tracker-Filter}(D^{low, c}_{t}, T_{t, t-1}),}
            \label{eq:tracker_filter}
        \end{equation}
        \textcolor{black}{where $D^{low}_{t}$ represents the low-score observations recalled.
        $T_{t, t-1}$ represents the predicted tracklets (\cref{Motion}).
        $\text{Tracker-Filter}(\cdot)$ is our proposed appearance filter, implemented by a \textbf{pseudo} visual tracker.
        We refer to this tracker as '\textbf{pseudo}' since it performs without tracklet lifecycle management.
        $D^{high}_{t}$ and $D^{low}_{t}$ sequentially conduct motion and appearance matching with existing tracklets $T_{t, t-1}$.}

        \textcolor{black}{\textbf{Geometry Filter.}}
        \label{geometry-filter}
        Inspired by~\cite{bevdet}, we introduce an elegant solution: we first employ category-specific factors $\theta_{s}$ to scale up $D^{r}_{t}$ size, followed by the standard NMS.
        Different from~\cite{bevdet}, the cross-object inter-frame similarity is characterized using BEV Intersection over Union ($IoU_{bev}$) and BEV Generalized IoU ($gIoU_{bev}$), enabling similarity assessment for non-overlapping objects and consider the distance similarity.        
        \textcolor{black}{As demonstrated in Tables \ref{table:ablation} and \ref{table:geometry}, this geometry filter efficiently removes numerous FP, especially for small object categories (\textit{Ped}, \textit{Bic}, etc.).
        A key insight is that size scaling and $gIoU_{bev}$ make predictions with similar 3D information interact effectively.}

        \textcolor{black}{\textbf{Pseudo Visual Tracker Filter.}}
        \label{tracker-filter}
        \textcolor{black}{To harness the potentially valuable appearance information from low-score detections, we utilize a visual tracker to recall these detections further.}
        This tracker firstly employs our proposed multi-camera appearance similarity metric MCAS (\cref{multi-camera metric}) to compute the visual cost between $D^{low, c}_{t}$ and $T_{t, t-1}$:
        \begin{equation}
            \textcolor{black}{C^{pre}_{mv} = -\text{MCAS}(D^{low, c}_{t},T_{t, t-1})},
            \label{eq:tracker_filter_mcas}
        \end{equation}
        \textcolor{black}{where $C^{pre}_{mv}$ ($mv$ for multi-view, $pre$ for pre-processing) represents the appearance cost between low-score detections and all alive tracklets.}
        Hungarian algorithm~\cite{hungarian} with hand-crafted thresholds $\theta_{tf}$ ($tf$ for tracker-filter) is then utilized to match $C^{pre}_{mv}$. 
        Matched detections $D^{low}_{t} \in D^{low, c}_{t}$ are recalled as potential valid visual observations, with the number of $N_{low}$.
        To avoid introducing 3D FP, we further implement a two-step verification process using matched tracklets $T^{pre}_{t-1} \in T_{t, t-1}$ in the subsequent association module. 
        Their noises are also heuristically modeled to characterize the reliability of their corresponding motion observations.
        The inner logic in this implementation is that the tracking system functions as a filter in 3D perception system since it removes background based on the tracklet-object affinity.
        

	\subsection{Motion-Appearance Data Association Module}
        \label{Association}
        Using a single multi-camera detector input, we perform associations in various embedding spaces to enhance 3D tracking performance and address the multi-modal alignment challenges.
        Motion matching first occurs with relatively reliable high-score detections in 3D space.
        Subsequently, appearance matching with two-step verification is executed to recall visually reliable low-score observations in the image space.
        Notably, we introduce a novel appearance metric to characterize inter-object similarity across multiple cameras.

        \label{multi-camera metric}
        \textbf{Multi-Camera Appearance Similarity Metric (MCAS).}
        Multi-view trackers typically construct appearance affinity implicitly, which is unsuited for TBD trackers.
        Besides, the prevalence of mono datasets (KITTI~\cite{kitti}, MOT20~\cite{mot20}, etc.) overlooks similarity representation in multi-camera setting.
        To address these, we propose a novel multi-view appearance similarity metric applicable across TBD trackers.
        \textcolor{black}{Specifically, given two 3D bounding boxes ($B_{1}^{3d}$ and $B_{2}^{3d}$) representing detection or/and tracklet, we first project them onto all surrounding cameras and aggregate the 2D states for each 3D object, which is formulated as:
        \begin{equation}
        \small
            B_{i, k}^{2d} = \text{Project}(B_{i}^{3d}, E_{k}, I_{k}),
            B_{i}^{2d} = \text{Concat}(B_{i, k}^{2d}|k=1, \dots,K) ,
            \label{concat_project}
        \end{equation}
        where $K$ is the number of cameras.
        $E_{k} \in \mathbb{R}^{3 \times 4}$ and $I_{k} \in \mathbb{R}^{3 \times 3}$ are the extrinsic and intrinsic matrices of the $k$-th camera.
        $\text{Project}(\cdot)$ is the 3D-to-2D project function.}
        $B_{i, k}^{2d} \in \mathbb{R}^{1 \times 4}$ represents the 2D bbox of $B_{i}^{3d}$ on the $k$-th camera, containing top-left and bottom-right pixel coordinates.
        \textcolor{black}{$\text{Concat}(\cdot)$ is the concatenate function along the camera.
        $B_{i}^{2d} \in \mathbb{R}^{K \times 4} $ is the overall 2D state of $B_{i}^{3d}$.
        Notably, the project coordinates are filled with invalid values in the camera that cannot capture $B_{i}^{3d}$.
        Next, we compute the similarity between $B_{1}^{3d}$ and $B_{2}^{3d}$ independently in each camera:
        \begin{equation}
            S_{ind} = \text{Concat}(\text{App}(B_{1, k}^{2d}, B_{2, k}^{2d})|k=1, \dots,K),
            \label{single_app}
        \end{equation}
        where $\text{App}(\cdot)$ is an off-the-shelf appearance distance, implemented in RockTrack via $IoU_{2d}$ or $gIoU_{2d}$ operations.
        $S_{ind}\in\mathbb{R}^{K \times 1 \times 1}$ ($ind$ for individual) is the similarity without view interaction.}
        Notably, in each camera, the affinity is valid only when this camera observes both $B_{1}^{3d}$ and $B_{2}^{3d}$.
        \textcolor{black}{We then perform explicit cross-camera aggregation on $S_{ind}$:
        \begin{equation}
            S_{mv} = \text{Fuse}(S_{ind}),
            \label{fuse_all_camera}
        \end{equation}
        where $\text{Fuse}(\cdot)$ is the fusion function, implemented in RockTrack via summation $\text{Sum}(\cdot)$, maximum $\text{Max}(\cdot)$, and average $\text{Avg}(\cdot)$ operations.
        $S_{mv} \in \mathbb{R}^{1 \times 1}$ ($mv$ for multi-view) represents the final multi-camera appearance similarity between $B_{1}^{3d}$ and $B_{2}^{3d}$.
        For clarity, Fig.~\ref{fig:f2}(b) visualizes this calculation process.
        The overall calculation process is abbreviated as:
        \begin{equation}
            S_{mv} = \text{MCAS}(B_{1}^{3d}, B_{2}^{3d}).
            \label{abbrev_mcas}
        \end{equation}}
        By extension, given two sets of 3D bounding boxes ($\beta_{1}$, $\beta_{2}$), the dimensions of $S_{mv}$ is $\mathbb{R}^{N_{1} \times N_{2}}$, where $N_{1}$ and $N_{2}$ are the number of elements in $\beta_{1}$ and $\beta_{2}$.
        MCAS offers two key advantages:
        (\uppercase\expandafter{\romannumeral1}): It is explicit, unsupervised, universally applicable to trackers, and robust to sensor failures, as proven in Table \ref{table:camera_drop}.
        (\uppercase\expandafter{\romannumeral2}): It is highly extensible, compatible with learning-based metrics and category-specific technology~\cite{polymot} through modifications of $\text{App}(\cdot)$ and $\text{Fuse}(\cdot)$.

        
        \textcolor{black}{\textbf{First Association}} occurs in 3D and BEV spaces.
        \label{moiton match}
        Based on $gIoU_{bev}$ and $gIoU_{3d}$, \textcolor{black}{a category-optimal cost matrix $C_{mo} \in \mathbb{R}^{N_{cls} \times N_{high} \times N_{tra}}$ ($mo$ for motion) is constructed between $D^{high}_{t}$ and $T_{t, t-1}$.
        $N_{cls}$, $N_{high}$, and $N_{tra}$ represent the number of categories, high-score detections $D^{high}_{t}$, and tracklets $T_{t, t-1}$, respectively.}
        Hungarian algorithm~\cite{hungarian} with category-optimal thresholds $\theta_{mo}$ is utilized to match $C_{mo}$, obtaining matched/unmatched detections $D^{1m}_{t}$/$D^{1u}_{t}$ and matched/unmatched tracklets $T^{1m}_{t-1}$/$T^{1u}_{t-1}$.
        $N^{um}_{high}$ and $N^{um}_{tra}$ denote the number of $D^{1u}_{t}$ and $T^{1u}_{t-1}$.
        
        \textcolor{black}{\textbf{Second Association}} is implemented in the image (2D) space to reduce FN matches and recall valid visual observations, by combining the low-score detection $D^{low}_{t}$.
        \label{appearance match}
        \textcolor{black}{Based on MCAS, the category-specific cost matrix $C_{app} \in \mathbb{R}^{N_{cls} \times (N_{low} + N^{um}_{high}) \times N^{um}_{tra}}$ ($app$ for appearance) is computed between $D^{1u}_{t}$, $D^{low}_{t}$ and $T^{1u}_{t-1}$.}
        Hungarian algorithm~\cite{hungarian} with thresholds $\theta _{app}$ are then applied to match $C_{app}$, obtaining matched detections/tracklets $D^{2m}_{t}$/$T^{2m}_{t-1}$ and final unmatched detections/tracklets $D^{um}_{t}$/$T^{um}_{t-1}$.
        Low/high-score detections in $D^{um}_{t}$ and $D^{2m}_{t}$ are regarded as $D^{um, low}_{t}$/$D^{um, high}_{t}$ and $D^{2m, low}_{t}$/$D^{2m, high}_{t}$, respectively.

        \textbf{Two-Step Verification.}
        Intuitively, TP observations should exhibit strong affinities with corresponding trajectories across distinct metrics, as also evidenced in Table~\ref{table:noise_factor}.
        Therefore, we assess the consistency of the corresponding tracklets of $D^{2m, low}_{t}$ in the association and pre-processing modules, i.e., $T^{2m}_{t-1}$ and $T^{pre}_{t-1}$.
        Detections failing verification are discarded.
        

	\subsection{Noise-Adaptive Motion Module}
        \label{Motion}

        \textbf{Predict.}
        We utilize \textit{Bicycle} and \textit{CTRA} motion models to perform the state transition (from $T_{t-1}$ to $T_{t,t-1}$) based on the Extended Kalman Filter (EKF).
        As introduced above, $T_{t, t-1}$ is utilized in the pre-processing and association.

        \textbf{Update.}
        \textcolor{black}{Most trackers employ a fixed~\cite{camomot, eagermot, deepfusionmot} or confidence-based~\cite{bytetrackv2} measurement noise.
        However, detections recalled through visual matching tend to be motion-noisier compared to geometry-based correlation, as also proven in Table~\ref{table:noise_factor}.}
        Consequently, we model the motion noise of recalled observations based on their matching modality and detection score. 
        The process is formulated as:
        \begin{equation}
            R^{i} = 10^{\alpha^{i}} \times (1 - c^{i})^{2} \times \mathbb{I}, 
            \label{model}
        \end{equation}
        where $R^{i}$ denotes measurement noise of the $i$-th matched detection $D^{m, i}_{t}$. 
        $\alpha^{i}$ and $c^{i}$ are the match stage index (0 for motion match, 1 for appear match) and confidence of $D^{m, i}_{t}$. 
        $\mathbb{I}$ is the identity matrix, with its dimension depending on the measurement capability of the detector.
        $R^{i}$ and $D^{m, i}_{t}$ are utilized to update the corresponding tracklet with the EKF.

	\subsection{Lifecycle Module}
        \label{Lifecycle}
        High-score unmatched detections $D^{um, high}_{t}$ are initialized as active tracklets, while low-score unmatched detections $D^{um, low}_{t}$ become tentative tracklets, representing occluded new objects. 
        Tentative tracklets transition to active status after $M_{hit}$ consecutive hits or are discarded otherwise. 
        Tracklets without hits for $M_{age}$ consecutive frames are eliminated. 
        This module merges new-born, matched, and unmatched-active tracklets into the final set of alive tracklets $T_{t}$. 
\begin{table}[t]
        \vspace{0.5em}
        \begin{center}
        \caption[l]{
        {Comparison of existing methods on the nuScenes test set. \textcolor{red}{\textbf{Red}}: Best. \textcolor{black}{\textbf{Black}}: Second best.}}
        \vspace{-0.6em}
        \label{table:test}
        \renewcommand{\arraystretch}{0.7}
        \setlength{\tabcolsep}{1.7mm}{
        \begin{tabular}{cccccc}
        \toprule
        \bf{Tracker}  & \bf{AMOTA$\uparrow$} & \bf{AMOTP$\downarrow$} & \bf{IDS$\downarrow$}  & \bf{FP$\downarrow$}   & \bf{FN$\downarrow$} \\ \midrule[0.8pt]
        DORT~\cite{dort}       & \textcolor{black}{\textbf{57.6}}  & \textcolor{black}{\textbf{0.951}}  & 774 & \textcolor{black}{\textbf{17049}} & 34660 \\
        ByteTrackv2~\cite{bytetrackv2}     & 56.4  & 1.005  & 704 & 18939 & \textbf{33531} \\
        DQTrack~\cite{li2023end}     & 52.3  & 1.096  & 1204 & 17279 & 36355 \\
        ADA-Track~\cite{ding2024ada} & 45.6 & 1.237 & 834 &  15699 & 39680 \\
        PFTrack~\cite{pftrack}    & 43.4  & 1.252 & \textcolor{red}{\textbf{249}} & 19048 & 42758 \\
        MoMA-M3T~\cite{huang2023delving} & 42.5 & 1.240 & 3136 & 18563 & 42934\\
        MUTR3D~\cite{mutr3d}    & 27.0  & 1.494 & 6018 & \textcolor{red}{\textbf{15372}} & 56874 \\\hline
        Ours       & \textcolor{red}{\textbf{59.1}}  & \textcolor{red}{\textbf{0.927}}  & \textcolor{black}{\textbf{630}} & 17774 & \textcolor{red}{\textbf{32695}} \\ \bottomrule
        \end{tabular}}
        \end{center}
        \vspace{-3em}
        \end{table} 

	\section{Experiments}
	\subsection{Datasets and Experimental Settings}
	\textbf{nuScenes.}
        nuScenes contains 1K scenes with 1.4M annotations. 
        Keyframes (2 Hz) are captured by 6 cameras and other sensors. 
        AMOTA~\cite{ab3dmot}, the primary metric, integrates FP, FN, and IDS across distinct recall rates.
        Evaluation covers 7 categories: \textit{Car}, \textit{Bic} (Bicycle), \textit{Moto} (Motorcycle), \textit{Ped} (Pedestrian), \textit{Bus}, \textit{Tra} (Trailer), and \textit{Tru} (Truck).

        \textbf{Baseline.}
        Poly-MOT~\cite{polymot}, without the second stage, is the baseline due to its robust performance in multi-category scenarios. 
        We finetune SF and first association thresholds for camera-only settings, as detailed in the following.

        \textbf{Implementation Details.}
        \label{details}
        RockTrack is implemented in Python using a 9940X CPU.
        HeightTrans~\cite{li2024dualbev} is used as the detector in the val and test sets.
        Hyperparameters yielding optimal AMOTA on the validation set are applied to the test set.
        category-specific $\theta_{sf}$ are: (\textit{Car}: 0.2; \textit{Ped}: 0.35; \textit{Moto}: 0.29; \textit{Tru}: 0.23; \textit{Tra}: 0.12; \textit{Bic}: 0.28; \textit{Bus}: 0.14).
        Scale factors $\theta_{s}$ are: (\textit{Bic}: 1.9; \textit{Moto}: 1.7; \textit{Ped}: 2.3), 1 for others.
        Standard NMS follows~\cite{polymot}, except $gIoU_{bev}$ for \textit{Ped}.
        In MCAS, $\text{App}(\cdot)$ are $IoU_{2D}$ and $gIoU_{2D}$ for pre-processing and association, $\text{Fuse}(\cdot)$ is $\text{Sum}(\cdot)$ for all modules.
        \textcolor{black}{The thresholds of visual tracker $\theta_{tf}$ are (\textit{Car}: -1.8; \textit{Ped, Tru}: -1.5; \textit{Moto, Tra}: -0.8; \textit{Bic, Bus}: -0.3)}.
        The thresholds of motion $\theta_{mo}$ and appearance $\theta_{app}$ associations are (\textit{Bic}: 1.6; \textit{Moto, Tra}: 1.5; \textit{Bus, Car, Tru}: 1.3; \textit{Ped}: 1.7) and \textcolor{black}{(\textit{Bus, Ped, Tra, Tru}: -3.8; \textit{Bic}: -0.9; \textit{Moto}: -1.4; \textit{Car}: -3.3), respectively.
        The similarity metric of motion matching, and lifecycle module setting are consistent with~\cite{polymot}.}
        We only utilize the keyframe for online tracking.
        Tracking latency is measured from input image to final tracking output, encompassing detection and tracking, ensuring fair comparison with end-to-end trackers.

        \begin{table}[t]
        \vspace{0.5em}
        \begin{center}
        \caption[l]{
        {Comparison of different trackers with the same detector on the nuScenes val set. 
        $\ddagger$: end-to-end tracker, generally demonstrating better detection accuracy.
        \textcolor{black}{\textbf{RockTrack uses consistent parameters across detectors.} }
        NDS considers only 7 tracking categories. 
        \cite{streampetr} is reproduced using \href{https://github.com/exiawsh/StreamPETR}{official code} due to unavailable tracking performance.
        }}
        \vspace{-0.6em}
        \label{table:val}
        \renewcommand{\arraystretch}{0.7}
        \setlength{\tabcolsep}{1mm}{
        \resizebox{\linewidth}{!}{
        \begin{tabular}{cccccccc}
        \toprule
        \multicolumn{2}{c}{\textbf{Detector}}                                                                                       & \multicolumn{6}{c}{\textbf{Tracker}}                                       \\ 
        \cmidrule(l{8.5mm}r{3.5mm}){1-2}\cmidrule(l{9.5mm}r{3.5mm}){3-8}
        \textbf{Method}                                      & \textbf{NDS$\uparrow$}                & \textbf{Method}     & \textbf{AMOTA$\uparrow$} & \textbf{AMOTP$\downarrow$} & \textbf{IDS$\downarrow$}  & \textbf{FP$\downarrow$}    & \textbf{FN$\downarrow$}  
        \\ \midrule[0.8pt]
        DETR3D$\ddagger$~\cite{DETR3D}        & 41.75$\ddagger$  & MUTR3D$\ddagger$~\cite{mutr3d}     & 29.4  & 1.498 & 3822 & 13321     & 46822 \\
        DETR3D$\ddagger$~\cite{DETR3D}       & 41.75$\ddagger$                          & PFTrack$\ddagger$~\cite{pftrack}    & 36.2  & 1.363 & 300  & \textbf{\text{--}}     & \textbf{\text{--}}\\ 
        DETR3D$\ddagger$~\cite{DETR3D}        & 41.75$\ddagger$  & DQTrack$\ddagger$~\cite{li2023end}     & 36.7  & 1.351 & 1120 & \textbf{\text{--}}     & \textbf{\text{--}} \\
        \textcolor{black}{DETR3D$\ddagger$~\cite{DETR3D}}        & \textcolor{black}{41.75$\ddagger$}  & \textcolor{black}{ADA-Track$\ddagger$~\cite{ding2024ada}}     & \textcolor{black}{37.8}  & \textcolor{black}{1.391} & \textcolor{black}{981} & \textcolor{black}{15443}     & \textcolor{black}{38466} \\
        \textcolor{black}{DETR3D~\cite{DETR3D}}        & \textcolor{black}{41.75}  & \textcolor{black}{CC-3DT~\cite{cc3dt}}     & \textcolor{black}{35.9}  & \textcolor{black}{1.361} & \textcolor{black}{2152} & \textcolor{black}{\textbf{\text{--}}}     & \textcolor{black}{\textbf{\text{--}}} \\
        \textcolor{black}{DETR3D~\cite{DETR3D}}        & \textcolor{black}{41.75}  & \textcolor{black}{MoMA-M3T~\cite{huang2023delving}}     & \textcolor{black}{36.2}  & \textcolor{black}{1.369} & \textcolor{black}{\textbf{\text{--}}} & \textcolor{black}{\textbf{\text{--}}}     & \textcolor{black}{\textbf{\text{--}}} \\
        \textcolor{black}{DETR3D~\cite{DETR3D}}        & \textcolor{black}{41.75}  & \textcolor{black}{Poly-MOT~\cite{polymot}}     & \textcolor{black}{37.7}  & \textcolor{black}{1.295} & \textcolor{black}{939} & \textcolor{black}{17330}     & \textcolor{black}{41096} \\
        \rowcolor{mypink} \textcolor{black}{DETR3D~\cite{DETR3D}}       & \textcolor{black}{41.75}                          & \textcolor{black}{\textbf{RockTrack}}   &  \textcolor{black}{38.1}                    &  \textcolor{black}{1.351}                     &  \textcolor{black}{698}                      & \textcolor{black}{14606} & \textcolor{black}{42416}   
                                   \\ 
        HeightTrans~\cite{li2024dualbev} & 53.35 & CBMOT~\cite{cbmot}       & 36.8 & 1.114 & 684 & 16212 & 41248 
        \\
        HeightTrans~\cite{li2024dualbev}  & 53.35                          & SimpleTrack~\cite{simpletrack} & 39.0   & 1.140    & 856  & 14749    & 43689    
        \\
        HeightTrans~\cite{li2024dualbev}  & 53.35                          & Poly-MOT~\cite{polymot}     & 46.1 & 1.147 & 521 & 14542 & 39904 
        \\
        \rowcolor{mypink} \textcolor{black}{HeightTrans~\cite{li2024dualbev}}  & \textcolor{black}{53.35}                          & \textcolor{black}{\textbf{RockTrack}}   & \textcolor{black}{48.8} & \textcolor{black}{1.148} &  \textcolor{black}{500}  &  \textcolor{black}{15106}  & \textcolor{black}{38688} 
        \\ 
        \textcolor{black}{StreamPETR~\cite{streampetr}}             & \textcolor{black}{57.88} & \textcolor{black}{StreamPETR~\cite{streampetr}} & \textcolor{black}{42.2} & \textcolor{black}{1.185} & \textcolor{black}{785} & \textcolor{black}{12537} & \textcolor{black}{41461} \\
            \rowcolor{mypink} \textcolor{black}{StreamPETR~\cite{streampetr}}             & \textcolor{black}{57.88} & \textcolor{black}{\textbf{RockTrack}} & \textcolor{black}{51.4} & \textcolor{black}{1.144} & \textcolor{black}{295} & \textcolor{black}{13971} & \textcolor{black}{34386} 
        \\ \bottomrule
        \end{tabular}}}
        \end{center}
        \vspace{-2.5em}
        \end{table}

	\subsection{Comparison with the State-of-the-Art Methods} 
	\textbf{Test.}
        \textbf{RockTrack ranks first on the nuScenes vision-only tracking leaderboard with 59.1\% AMOTA} with HeightTrans~\cite{li2024dualbev} detector\footnote{To be fair, we do not compare trackers with the ViT-Large~\cite{dosovitskiy2020image}, as AMOTA heavily depends on detector performance.}.
        Table~\ref{table:test} illustrates RockTrack superiority over all learning-based trackers~\cite{mutr3d, pftrack, dort, li2023end, huang2023delving}.
        Compared to filter-based methods, RockTrack outperforms the closest competitor ByteTrackv2~\cite{bytetrackv2} with +2.7\% AMOTA, -74 IDS.
        Besides, RockTrack achieves the second-lowest IDS 630 and the second-lowest FN 32695, emphasizing its stable tracking without compromising recall.
 
        \textbf{Val.}
        As illustrated in Table~\ref{table:val}, we perform a thorough and fair comparison with other advanced methods using the same detector.
        With the same HeightTrans~\cite{li2024dualbev} detector, RockTrack outperforms competing general (SimpleTrack~\cite{simpletrack}, Poly-MOT~\cite{polymot}, CBMOT~\cite{cbmot}) trackers by a significant margin.
        The parameter complexity of the TBD framework is one of its primary limitations.
        Therefore, we generalize the optimal hyperparameters (details in \cref{details}) to other public detectors.
        RockTrack surpasses the existing camera-specific (CC-3DT~\cite{cc3dt}) trackers.
        Compared to end-to-end camera-only trackers~\cite{pftrack, mutr3d, li2023end, ding2024ada} utilizing the same DETR3D~\cite{DETR3D} head, RockTrack incurs minimal computational overhead (CPU only) while delivering superior performance \textcolor{black}{(+1.9\% AMOTA to PFTrack~\cite{pftrack}, +1.4\% AMOTA to DQTrack~\cite{li2023end}, +0.4\% AMOTA to ADA-Track~\cite{ding2024ada}}.
        When paired with a more powerful detector~\cite{streampetr}, RockTrack achieves comparable results with LiDAR-based trackers.
        \textcolor{black}{In summary, following the TBD framework, RockTrack demonstrates impressive scalability and accuracy with various detectors without requiring training or fine-tuning.}

        \begin{table}[t]

        \centering
        \caption[l]{
        {Ablation studies of each module on the nuScenes val set.
        \textbf{Pre:} Pre-processing.
        \textbf{Est:} Motion estimation.
        \textbf{Asso:} Association.
        \textcolor{black}{\textbf{G:} Geometry filter integration.
        \textbf{P:} Pseudo visual tracker for low-confidence detection recall.
        \textbf{H:} Heuristic noise modeling.
        \textbf{S:} Secondary appearance-based matching.}}}
        \vspace{-0.4em}
        \label{table:ablation}
        \renewcommand{\arraystretch}{0.5}
        \setlength{\tabcolsep}{2mm}{
        \begin{tabular}{c|cccc|ccc}
        \toprule
        \multirow{2.5}{*}{\textbf{Index}} & \multicolumn{2}{c}{\textbf{Pre}} & \textbf{Est} & \textbf{Asso}  & \multirow{2.5}{*}{\textbf{AMOTA$\uparrow$}} & \multirow{2.5}{*}{\textcolor{black}{\textbf{MOTA$\uparrow$}}} & \multirow{2.5}{*}{\textbf{IDS$\downarrow$}} \\ \cmidrule(l{2.8mm}r{2.8mm}){2-3} \cmidrule(l{2.7mm}r{2.7mm}){4-4} \cmidrule(l{2.9mm}r{2.6mm}){5-5}
                                        & \textbf{G}   & \textbf{P}   & \textbf{H}  & \textbf{S} &                                 &                                 &                                                                                       \\ \midrule[0.8pt]
        Baseline  &  \textbf{\text{--}}                        & \textbf{\text{--}}       & \textbf{\text{--}}                & \textbf{\text{--}}                          &  46.1              &   \textcolor{black}{38.4}           &  \textcolor{black}{546}                                    \\
        EXP1      & \checkmark        & \textbf{\text{--}}          & \textbf{\text{--}}        & \textbf{\text{--}}                          &  47.6              &   \textcolor{black}{40.1}           &  \textcolor{black}{503}                                  \\
        EXP2      & \textbf{\text{--}}                          & \textbf{\text{--}}         & \checkmark                & \textbf{\text{--}} &              46.7            & \textcolor{black}{39.2}            &  \textcolor{black}{\textbf{491}}                                  \\
        EXP3      & \checkmark & \textbf{\text{--}}        & \checkmark                  & \textbf{\text{--}} &     48.1           &   \textcolor{black}{40.8}          &   \textcolor{black}{497}                                     \\
        EXP4      & \checkmark & \textbf{\text{--}}        & \checkmark                  & \checkmark &   48.1             & \textcolor{black}{\textcolor{black}{40.7}}              &  \textcolor{black}{\textcolor{black}{492}}          \\
        EXP5      & \checkmark & \checkmark & \checkmark & \checkmark &   \textcolor{black}{\textbf{48.8}}             & \textcolor{black}{\textbf{41.1}}              &  \textcolor{black}{500}                    
         \\ \bottomrule
        \end{tabular}}
        \vspace{-2em}
        \end{table}

        \begin{table}[t]
        \vspace{0.5em}
        \begin{center}
        \caption[l]{
        {Ablation studies with different geometry filters in the pre-processing module on the nuScenes val set. 
        \textcolor{black}{\textbf{Scale}: objects are scaled up, followed by classic-NMS.}
        \textbf{Delete Dets}: The number of eliminated high-score detections.}}
        \vspace{-0.4em}
        \label{table:geometry}
        \renewcommand{\arraystretch}{0.5}
        \setlength{\tabcolsep}{1.5mm}{
        \begin{tabular}{cccccc}
\toprule
\bf{Category}              & \bf{Scale}  & \bf{Metric}          & \bf{AMOTA$\uparrow$}                & \bf{IDS$\downarrow$}                  & \bf{Delete Dets}                     \\ \hline
\multirow{3}{*}{\textit{Ped}}  & \xmark ($\theta_{s}$=1) & $IoU_{bev}$          &  42.5                    &   345                   &       0                                \\
                      & \checkmark ($\theta_{s}$=2.3) & $IoU_{bev}$  & 44.1                     & \textbf{296}                     &    12124                                 \\
                      & \checkmark ($\theta_{s}$=2.3) & $gIoU_{bev}$ &  \textbf{46.7}                    &   311                   &   7099                                \\\hline
\multirow{2}{*}{\textit{Bic}}  & \xmark ($\theta_{s}$=1) & $IoU_{bev}$          &  42.3                    &   6                   &     590                         \\
                      & \checkmark ($\theta_{s}$=1.9) & $IoU_{bev}$  &   \textbf{45.2}                   &   \textbf{1}                   &   1101                             \\ \hline
\multirow{2}{*}{\textit{\textcolor{black}{Car}}} & \textcolor{black}{\xmark ($\theta_{s}$=1)} & \textcolor{black}{$IoU_{bev}$}         &  \textcolor{black}{\textbf{65.7}}                    &   \textcolor{black}{176}                               &      \textcolor{black}{22224}  \\
                              & \textcolor{black}{\checkmark ($\theta_{s}$=1.1)} & \textcolor{black}{$IoU_{bev}$}  &  \textcolor{black}{65.5}                    &   \textcolor{black}{\textbf{130}}                    &  \textcolor{black}{27072}               \\
                              \bottomrule
        \end{tabular}}
        \end{center}
        \vspace{-2.3em}
        \end{table} 

        \subsection{Run-Time Discussion}
        On a constrained platform (9940X CPU), RockTrack achieves superior latency performance (3.8  FPS (5 FPS without detection)) on nuScenes among advanced learning-base/free methods (PFTrack~\cite{pftrack} 2 FPS on A100,
        DQTrack~\cite{li2023end} 4.8 FPS on A100,
        \textcolor{black}{CC-3DT~\cite{cc3dt} 2.1 FPS on RTX 3090}, 
        ADA-Track~\cite{ding2024ada} 3.3 FPS on 2080Ti).
        We also analyze the time consumption for appearance-related operations.
        Benefiting from the parallel nature of MCAS, the latencies for the tracker-filter and the second appearance association are both 1.6 ms.
        However, 3D-2D projections introduce notable overhead (42 ms).
        Nevertheless, as shown in Table~\ref{table:ablation}, the accuracy gains justify these costs, and the overall latency (5 FPS) remains competitive.


	\subsection{Ablation Experiment}

        We conduct extensive ablation experiments of each module on the nuScenes val set, as illustrated in Table~\ref{table:ablation}.
        HeightTrans~\cite{li2024dualbev} is utilized as the detector for the ablation study.
 
	\textbf{The Effect of the Geometry Filter. }
        The geometry filter (EXP1) led to an improvement in AMOTA by +1.5\% and MOTA by +1.7\%.
        Table \ref{table:geometry} reveals the reason for the improvement: the geometry filter substantially reduces FP in detections, particularly for small categories (\textit{Ped, etc.}), as evidenced by the significant increase in the number of deleted detections.
        Besides, $gIoU_{bev}$ outperforms $IoU_{bev}$ on \textit{Ped} (+2.6\% AMOTA) by considering location-distance similarity more comprehensively.
        \textcolor{black}{The limitation on large objects (\textit{Car}, etc.) is also noteworthy (-0.2\% AMOTA).
        Since the 3D information of these categories is relatively accurate, the geometry filter can lead to recall loss.}


        \begin{table}[t]
        \begin{center}
        \caption[l]{
        {\textcolor{black}{Ablation studies of introducing stage factor $\alpha$ to model the reliability of observations (\textbf{Left}) and introducing two-step verification (\textbf{Right}) on the nuScenes val set.}}}
        \label{table:noise_factor}
        \renewcommand{\arraystretch}{0.7}
        \vspace{-0.5em}
        \setlength{\tabcolsep}{2mm}{
        \resizebox{\linewidth}{!}{
        \begin{tabular}{c|c|c|c||c|c|c|c}
        \toprule
        \bf{\textcolor{black}{Model}} &\textcolor{black}{\bf{AMOTA$\uparrow$}}   & \textcolor{black}{\bf{MOTA$\uparrow$}} & \textcolor{black}{\bf{IDS$\downarrow$}}
        & \textbf{Verify} & \bf{AMOTA$\uparrow$} & \bf{MOTA$\uparrow$} & \bf{IDS$\downarrow$}
        \\ \midrule[0.8pt]
        \textcolor{black}{\xmark}       & \textcolor{black}{48.5}          &   \textcolor{black}{40.8}         & \textcolor{black}{529} & \textcolor{black}{\xmark} & 48.5 & 40.8 & 501\\
        \textcolor{black}{\checkmark}   & \textcolor{black}{\textbf{48.8}} &   \textcolor{black}{\textbf{41.1}} & \textcolor{black}{\textbf{500}} & \textcolor{black}{\checkmark} & \textbf{48.8 }& \textbf{41.1} & \textbf{500} \\ \bottomrule
        \end{tabular}}}
        \end{center}
        \vspace{-2em}
        \end{table}

        \textbf{The Effect of the Adaptive Noise.}
        The heuristic noise (EXP2) enhances tracker stability and robustness, resulting in a significant improvement in MOTA (+0.8\%) and AMOTA (+0.6\%), along with a drop in IDS (-55).
        \textcolor{black}{As illustrated in Table~\ref{table:noise_factor}, compared with treating all association stages equally (ByteTrackv2~\cite{bytetrackv2}), the introduction of stage factors improves the primary metrics (+0.3\% AMOTA and +0.3\% MOTA), confirming our inference that the motion reliability of recalled observations in different matching stages varies.}


        \begin{table}[t]
        \vspace{0.5em}
        \centering
        \caption[l]{
        {\textcolor{black}{Ablation studies with different treatments for coarse low-score detections $D^{low, c}_{t}$ on the nuScenes val set.
        EXP0 and EXP1 are EXP3 and EXP4 in Table~\ref{table:ablation}.
        EXP2 adopts the BYTE~\cite{bytetrackv2, simpletrack} approach, directly matching these detections with existing tracklets.
        EXP5 is our treatment, which first filters and then matches these detections.
        \textbf{How To Match}: the modality of matching low-score detections in the second association module.
        $IoU_{bev}$ is the motion metric.
        }}}
        \vspace{-2em}
        \label{table:two_stage}
        \renewcommand{\arraystretch}{0.5}
        \setlength{\tabcolsep}{0.8mm}{
        \begin{tabular}{c|ccccccc}
        \toprule
        \multirow{2.5}{*}{\textbf{\textcolor{black}{Method}}} & \multicolumn{2}{c}{\textbf{\textcolor{black}{How To Filter}}} & \multicolumn{2}{c}{\textbf{\textcolor{black}{How To Match}}}  & \multirow{2.5}{*}{\textbf{\textcolor{black}{AMOTA$\uparrow$}}} & \multirow{2.5}{*}{\textbf{\textcolor{black}{MOTA$\uparrow$}}}  \\ \cmidrule(l{2.8mm}r{2.8mm}){2-3} \cmidrule(l{2.7mm}r{2.7mm}){4-5}
        & \textcolor{black}{\textbf{Track-Filter}}   & \textcolor{black}{\textbf{No-Filter}}   & \textcolor{black}{\textbf{Appear}}  & \textcolor{black}{\textbf{Motion}}                         &                                         \\ \midrule[0.8pt]
        \textcolor{black}{EXP0}  &  \textbf{\text{--}}                        & \textbf{\text{--}}       & \textbf{\text{--}}                & \textbf{\text{--}}                                     &  \textcolor{black}{48.1}           &  \textcolor{black}{40.8}                  \\
        \textcolor{black}{EXP1}  &  \textbf{\text{--}}                        & \textbf{\text{--}}       & \textbf{\checkmark}                & \textbf{\text{--}}                                     &  \textcolor{black}{48.1}           &  \textcolor{black}{40.7}                  \\
        \textcolor{black}{EXP2}      & \textbf{\text{--}}        & \checkmark                & \textbf{\text{--}}      & \checkmark                             &  \textcolor{black}{48.3}           &  \textcolor{black}{40.9}                             \\
        \textcolor{black}{EXP3}      & \textbf{\text{--}}                          & \checkmark         & \checkmark                & \textbf{\text{--}}            &  \textcolor{black}{48.2}           & \textcolor{black}{40.6}                    \\
        
        \textcolor{black}{EXP4}      & \checkmark & \textbf{\text{--}}        & \textbf{\text{--}}                  & \checkmark          &   \textcolor{black}{48.6}          &    \textcolor{black}{40.9}                    \\
        
        \textcolor{black}{EXP5}      & \checkmark & \textbf{\text{--}} & \checkmark & \textbf{\text{--}}              &  \textcolor{black}{48.8}           &  \textcolor{black}{41.1}              
         \\ \bottomrule
        \end{tabular}}
        \vspace{-1.6em}
        \end{table}

	\textbf{The Effect of the Association Module and Tracker Filter. }
        As shown in Table~\ref{table:ablation}, solely utilizing high-score detections for the appearance association (EXP4) results in a slight decrease in MOTA (-0.1\%) compared to EXP3.
        A key insight is that detections in a single range but not associated through motion are more likely to represent 3D FP or observations of new tracklets.
        Therefore, to leverage the visual richness of multi-camera detectors, we incorporate low-score detections (EXP5) to obtain more informative measurements. 
        The implementation of the pseudo visual tracker (EXP5) and multi-stage association to recall and match low-score observations enhances AMOTA by 0.7\% and MOTA by 0.4\% compared to EXP4, demonstrating the efficacy of our detection recall strategy.
        Further discussion of our association and recall strategy follows.
        \vspace{-1.3em}

        \begin{table}[t]
        \begin{center}
        \caption[l]{
        {\textcolor{black}{Ablation studies of integrating different numbers of cameras on the nuScenes val set.
        The back, back-right, and back-left cameras are dropped in sequence.
        Alternate: back, front-right, and front-left cameras are removed, eliminating view overlapping among the remaining cameras.}}}
        \label{table:camera_drop}
        \renewcommand{\arraystretch}{0.5}
        \vspace{-0.4em}
        \setlength{\tabcolsep}{2.5mm}{
        \begin{tabular}{cccc}
        \toprule
        \bf{\textcolor{black}{Cam Num}} &\textcolor{black}{\bf{AMOTA$\uparrow$}}   & \textcolor{black}{\bf{MOTA$\uparrow$}} & \textcolor{black}{\bf{AMOTP$\downarrow$}} \\ \midrule[0.8pt]
        \textcolor{black}{3 (Alternate)}       & \textcolor{black}{48.4}          &   \textcolor{black}{40.7}     & \textcolor{black}{1.153}  \\
        \textcolor{black}{3}       & \textcolor{black}{48.5}          &   \textcolor{black}{40.8}     & \textcolor{black}{1.157}  \\
        \textcolor{black}{4}       & \textcolor{black}{48.5}          &   \textcolor{black}{41.0}     & \textcolor{black}{1.157}  \\
        \textcolor{black}{5}       & \textcolor{black}{48.5}          &   \textcolor{black}{40.8}     & \textcolor{black}{1.157}  \\
        \textcolor{black}{6}   & \textcolor{black}{48.8} &   \textcolor{black}{41.1} & \textcolor{black}{1.148}  \\ \bottomrule
        \end{tabular}}
        \end{center}
        \vspace{-3em}
        \end{table}

        \begin{figure*}[t]
          \centering
          \includegraphics[width=1\linewidth]{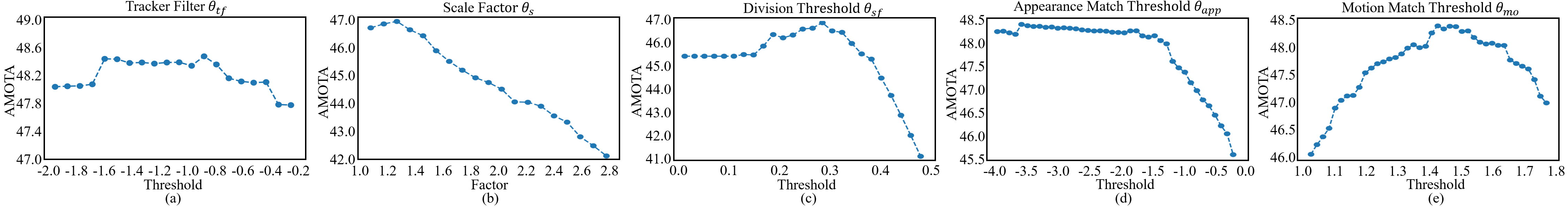}
          \caption[l]{\textcolor{black}{The comparison of the accuracy under distinct hyperparameter. 
          All categories are applied to the same parameter.
           }}
        \label{fig:f5}
        \vspace{-2em}
       \end{figure*}

        \textcolor{black}{\subsection{Matching Strategy Analysis}}
        \label{Matching Strategy Analysis}
        \textcolor{black}{In Table \ref{table:two_stage}, we conduct comparative experiments on distinct recall and association architectures.}
        
        \textcolor{black}{\textbf{Filter vs Not Filter.}
        Compared with BYTE (EXP2), our method (EXP5) demonstrates improvements of +0.5\% in AMOTA and +0.2\% in MOTA.
        Our tracker-filter enhances both similarity metrics due to improved input quality (EXP4 vs EXP2, EXP5 vs EXP3).
        It yields greater benefits for the appearance setting (+0.6\% AMOTA) than the motion setting (+0.3\% AMOTA).
        This disparity arises from the higher eagerness of visual reliability over motion reliability, as no-depth appearance matching is more ambiguous.}

        \textcolor{black}{\textbf{Appearance vs Motion.}
        Retaining all low-score detections (EXP2 and EXP3) yields superior motion metric performance (+0.1\% AMOTA), as visual association proves less robust than motion association. 
        However, when selectively recalling these detections (EXP4 and EXP5), the appearance metric dominates. 
        Utilizing reliable visual information from camera-only detections demonstrates greater potential, mitigating motion mismatches caused by imprecise spatial information in low-score detections.}

        \textcolor{black}{\textbf{Verify vs Not Verify.}
        Table \ref{table:noise_factor} demonstrates the criticality of two-step verification in enhancing the visual recall of low-score detections (+0.3\% AMOTA and +0.3\% MOTA).
        A key insight is that TP observations should display strong affinities with corresponding tracklets across different metrics.}
        \vspace{-1em}


        \textcolor{black}{\subsection{Robustness Analysis}}

        \textcolor{black}{\textbf{The Robustness of the Cameras Drops.}
        Table \ref{table:camera_drop} demonstrates the robustness of RockTrack to sensor failure, with tracking accuracy only decreasing by 0.4\% AMOTA even when half of the sensors failed.
        Benefiting from the explicit computing framework, MCAS can be integrated across diverse camera configurations.}


        \textcolor{black}{\textbf{The Insensitivity of the Hyperparameters.}
        Fig \ref{fig:f5} depicts the sensitivity of crucial hyperparameters in RockTrack, including two fundamental thresholds ($\theta_{sf}$ and $\theta_{mo}$) and three newly introduced parameters ($\theta_{s}$, $\theta_{tf}$, and $\theta_{app}$).
        $\theta_{tf}$ and $\theta_{app}$, which control low-score detection recall, exhibit insensitivity to adjustments.
        $\theta_{tf}$ has a clear edge between strict and reasonable thresholds and maintains a flat performance curve over a significant interval.
        Similarly, $\theta_{app}$ sustains stable improvement over a long range.
        Scale factor $\theta_{s}$ demonstrates effectiveness at small values, as excessive amplification brings FN.
        As with other TBD methods, $\theta_{sf}$ and $\theta_{mo}$ are category- and detector-specific, requiring careful calibration. 
        $\theta_{sf}$ curve also shows the high FP rate inherent to camera-only detectors, necessitating stringent thresholds for optimal performance.}

\section{CONCLUSIONS}

This paper proposes a robust 3D MOT method following the TBD framework specifically for multi-camera detectors, terms RockTrack.
RockTrack enhances tracking performance in multi-view settings by addressing the ill-posed nature and leveraging rich information from various representation spaces.
RockTrack achieves SOTA on the nuScenes vision-only tracking leaderboard. 
Without training and fine-tuning, RockTrack demonstrates superior performance across various detectors.
We anticipate RockTrack to be pivotal in multi-camera technology on practical platforms.









\bibliographystyle{IEEEtran}
\bibliography{IEEEabrv,myref.bib}

\begin{thebibliography}{10}
\providecommand{\url}[1]{#1}
\csname url@samestyle\endcsname
\providecommand{\newblock}{\relax}
\providecommand{\bibinfo}[2]{#2}
\providecommand{\BIBentrySTDinterwordspacing}{\spaceskip=0pt\relax}
\providecommand{\BIBentryALTinterwordstretchfactor}{4}
\providecommand{\BIBentryALTinterwordspacing}{\spaceskip=\fontdimen2\font plus
\BIBentryALTinterwordstretchfactor\fontdimen3\font minus \fontdimen4\font\relax}
\providecommand{\BIBforeignlanguage}[2]{{%
\expandafter\ifx\csname l@#1\endcsname\relax
\typeout{** WARNING: IEEEtran.bst: No hyphenation pattern has been}%
\typeout{** loaded for the language `#1'. Using the pattern for}%
\typeout{** the default language instead.}%
\else
\language=\csname l@#1\endcsname
\fi
#2}}
\providecommand{\BIBdecl}{\relax}
\BIBdecl

\bibitem{ab3dmot}
X.~Weng, J.~Wang, D.~Held, and K.~Kitani, ``Ab3dmot: A baseline for 3d multi-object tracking and new evaluation metrics,'' \emph{arXiv preprint arXiv:2008.08063}, 2020.

\bibitem{polymot}
X.~Li, T.~Xie, D.~Liu, J.~Gao, K.~Dai, Z.~Jiang, L.~Zhao, and K.~Wang, ``Poly-mot: A polyhedral framework for 3d multi-object tracking,'' \emph{arXiv preprint arXiv:2307.16675}, 2023.

\bibitem{centerpoint}
T.~Yin, X.~Zhou, and P.~Krahenbuhl, ``Center-based 3d object detection and tracking,'' in \emph{CVPR}, 2021, pp. 11\,784--11\,793.

\bibitem{simpletrack}
Z.~Pang, Z.~Li, and N.~Wang, ``Simpletrack: Understanding and rethinking 3d multi-object tracking,'' \emph{arXiv preprint arXiv:2111.09621}, 2021.

\bibitem{camomot}
L.~Wang, X.~Zhang, W.~Qin, X.~Li, J.~Gao, L.~Yang, Z.~Li, J.~Li, L.~Zhu, H.~Wang \emph{et~al.}, ``Camo-mot: Combined appearance-motion optimization for 3d multi-object tracking with camera-lidar fusion,'' \emph{TITS}, 2023.

\bibitem{eagermot}
A.~Kim, A.~O{\v{s}}ep, and L.~Leal-Taix{\'e}, ``Eagermot: 3d multi-object tracking via sensor fusion,'' in \emph{ICRA}.\hskip 1em plus 0.5em minus 0.4em\relax IEEE, 2021, pp. 11\,315--11\,321.

\bibitem{shasta}
T.~Sadjadpour, J.~Li, R.~Ambrus, and J.~Bohg, ``Shasta: Modeling shape and spatio-temporal affinities for 3d multi-object tracking,'' \emph{RA-L}, 2023.

\bibitem{deepfusionmot}
X.~Wang, C.~Fu, Z.~Li, Y.~Lai, and J.~He, ``Deepfusionmot: A 3d multi-object tracking framework based on camera-lidar fusion with deep association,'' \emph{RA-L}, vol.~7, no.~3, pp. 8260--8267, 2022.

\bibitem{cc3dt}
T.~Fischer, Y.-H. Yang, S.~Kumar, M.~Sun, and F.~Yu, ``Cc-3dt: Panoramic 3d object tracking via cross-camera fusion,'' 2022.

\bibitem{Monotrack}
H.-N. Hu, Y.-H. Yang, T.~Fischer, T.~Darrell, F.~Yu, and M.~Sun, ``Monocular quasi-dense 3d object tracking,'' \emph{IEEE TPAMI}, vol.~45, no.~2, pp. 1992--2008, 2023.

\bibitem{mutr3d}
T.~Zhang, X.~Chen, Y.~Wang, Y.~Wang, and H.~Zhao, ``Mutr3d: A multi-camera tracking framework via 3d-to-2d queries,'' in \emph{CVPR}, 2022, pp. 4537--4546.

\bibitem{li2024fastpoly}
X.~Li, D.~Liu, L.~Zhao, Y.~Wu, X.~Wu, and J.~Gao, ``Fast-poly: A fast polyhedral framework for 3d multi-object tracking,'' 2024.

\bibitem{ding2024ada}
S.~Ding, L.~Schneider, M.~Cordts, and J.~Gall, ``Ada-track: End-to-end multi-camera 3d multi-object tracking with alternating detection and association,'' in \emph{CVPR}, 2024, pp. 15\,184--15\,194.

\bibitem{lin2023sparse4d}
X.~Lin, Z.~Pei, T.~Lin, L.~Huang, and Z.~Su, ``Sparse4d v3: Advancing end-to-end 3d detection and tracking,'' \emph{arXiv preprint arXiv:2311.11722}, 2023.

\bibitem{lss}
J.~Philion and S.~Fidler, ``Lift, splat, shoot: Encoding images from arbitrary camera rigs by implicitly unprojecting to 3d,'' in \emph{ECCV}.\hskip 1em plus 0.5em minus 0.4em\relax Springer, 2020, pp. 194--210.

\bibitem{bevdet}
J.~Huang, G.~Huang, Z.~Zhu, Y.~Ye, and D.~Du, ``Bevdet: High-performance multi-camera 3d object detection in bird-eye-view,'' 2022.

\bibitem{DETR3D}
Y.~Wang, V.~C. Guizilini, T.~Zhang, Y.~Wang, H.~Zhao, and J.~Solomon, ``Detr3d: 3d object detection from multi-view images via 3d-to-2d queries,'' in \emph{CoRL}.\hskip 1em plus 0.5em minus 0.4em\relax PMLR, 2022, pp. 180--191.

\bibitem{BEVFormer}
Z.~Li, W.~Wang, H.~Li, E.~Xie, C.~Sima, T.~Lu, Y.~Qiao, and J.~Dai, ``Bevformer: Learning bird’s-eye-view representation from multi-camera images via spatiotemporal transformers,'' in \emph{ECCV}, 2022, pp. 1--18.

\bibitem{PETR}
Y.~Liu, T.~Wang, X.~Zhang, and J.~Sun, ``Petr: Position embedding transformation for multi-view 3d object detection,'' in \emph{ECCV}.\hskip 1em plus 0.5em minus 0.4em\relax Springer, 2022, pp. 531--548.

\bibitem{streampetr}
S.~Wang, Y.~Liu, T.~Wang, Y.~Li, and X.~Zhang, ``Exploring object-centric temporal modeling for efficient multi-view 3d object detection,'' \emph{arXiv preprint arXiv:2303.11926}, 2023.

\bibitem{pftrack}
Z.~Pang, J.~Li, P.~Tokmakov, D.~Chen, S.~Zagoruyko, and Y.-X. Wang, ``Standing between past and future: Spatio-temporal modeling for multi-camera 3d multi-object tracking,'' in \emph{CVPR}, 2023, pp. 17\,928--17\,938.

\bibitem{li2023end}
Y.~Li, Z.~Yu, J.~Philion, A.~Anandkumar, S.~Fidler, J.~Jia, and J.~Alvarez, ``End-to-end 3d tracking with decoupled queries,'' in \emph{ICCV}, 2023, pp. 18\,302--18\,311.

\bibitem{li2024dualbev}
P.~Li, W.~Shen, Q.~Huang, and D.~Cui, ``Dualbev: Cnn is all you need in view transformation,'' \emph{arXiv preprint arXiv:2403.05402}, 2024.

\bibitem{solofusion}
J.~Park, C.~Xu, S.~Yang, K.~Keutzer, K.~Kitani, M.~Tomizuka, and W.~Zhan, ``Time will tell: New outlooks and a baseline for temporal multi-view 3d object detection,'' 2023.

\bibitem{fcos3d}
T.~Wang, X.~Zhu, J.~Pang, and D.~Lin, ``Fcos3d: Fully convolutional one-stage monocular 3d object detection,'' in \emph{ICCV}, 2021, pp. 913--922.

\bibitem{huang2023delving}
K.-C. Huang, M.-H. Yang, and Y.-H. Tsai, ``Delving into motion-aware matching for monocular 3d object tracking,'' in \emph{ICCV}, 2023, pp. 6909--6918.

\bibitem{bytetrack}
Y.~Zhang, P.~Sun, Y.~Jiang, D.~Yu, F.~Weng, Z.~Yuan, P.~Luo, W.~Liu, and X.~Wang, ``Bytetrack: Multi-object tracking by associating every detection box,'' 2022.

\bibitem{hungarian}
H.~W. Kuhn, ``The hungarian method for the assignment problem,'' \emph{Naval research logistics quarterly}, vol.~2, no. 1-2, pp. 83--97, 1955.

\bibitem{kitti}
A.~Geiger, P.~Lenz, and R.~Urtasun, ``Are we ready for autonomous driving? the kitti vision benchmark suite,'' in \emph{CVPR}.\hskip 1em plus 0.5em minus 0.4em\relax IEEE, 2012, pp. 3354--3361.

\bibitem{mot20}
P.~Dendorfer, H.~Rezatofighi, A.~Milan, J.~Shi, D.~Cremers, I.~Reid, S.~Roth, K.~Schindler, and L.~Leal-Taix{\'e}, ``Mot20: A benchmark for multi object tracking in crowded scenes,'' \emph{arXiv preprint arXiv:2003.09003}, 2020.

\bibitem{bytetrackv2}
Y.~Zhang, X.~Wang, X.~Ye, W.~Zhang, J.~Lu, X.~Tan, E.~Ding, P.~Sun, and J.~Wang, ``Bytetrackv2: 2d and 3d multi-object tracking by associating every detection box,'' 2023.

\bibitem{dort}
Q.~Lian, T.~Wang, D.~Lin, and J.~Pang, ``Dort: Modeling dynamic objects in recurrent for multi-camera 3d object detection and tracking,'' 2023.

\bibitem{cbmot}
N.~Benbarka, J.~Schr{\"o}der, and A.~Zell, ``Score refinement for confidence-based 3d multi-object tracking,'' in \emph{IROS}.\hskip 1em plus 0.5em minus 0.4em\relax IEEE, 2021, pp. 8083--8090.

\bibitem{dosovitskiy2020image}
A.~Dosovitskiy, L.~Beyer, A.~Kolesnikov, D.~Weissenborn, X.~Zhai, T.~Unterthiner, M.~Dehghani, M.~Minderer, G.~Heigold, S.~Gelly \emph{et~al.}, ``An image is worth 16x16 words: Transformers for image recognition at scale,'' \emph{arXiv preprint arXiv:2010.11929}, 2020.

\end{thebibliography}

\end{document}